\pdfoutput=1

\documentclass[11pt]{article}

\usepackage{acl}

\usepackage{times}
\usepackage{latexsym}

\usepackage[T1]{fontenc}

\usepackage[utf8]{inputenc}

\usepackage{microtype}

\usepackage{inconsolata}

\usepackage{booktabs}
\usepackage{graphicx}
\usepackage{amsmath}
\usepackage{cleveref}
\usepackage{enumitem}
\usepackage{bbding}
\setlist{nosep}
\setenumerate{leftmargin=*}
\usepackage[noorphans,vskip=0.5ex,leftmargin=1em]{quoting}

\usepackage{wasysym}
\DeclareUnicodeCharacter{266B}{\twonotes}

\usepackage[norelsize,ruled,vlined,linesnumbered,noend]{algorithm2e}\DontPrintSemicolon
\crefname{algocf}{algorithm}{algorithms}
\Crefname{algocf}{Algorithm}{Algorithms}
\crefname{algocfline}{line}{lines}
\Crefname{algocfline}{Line}{Lines}

\title{Semi-Structured Chain-of-Thought: Integrating Multiple Sources of Knowledge for Improved Language Model Reasoning}

\author{Xin Su$^{1, 2}$  \quad 
        {Tiep Le$^{2}$} \quad
        Steven Bethard$^{1}$ \quad
        Phillip Howard$^{2}$ \quad \\
        $^{1}$University of Arizona \quad $^{2}$Intel Labs \\
        {\tt {\{xinsu, bethard\}}@arizona.edu}, \tt{\{tiep.le, phillip.r.howard\}}@intel.com}

\begin{document}
\maketitle

\begin{abstract}
An important open question in the use of large language models for knowledge-intensive tasks is how to effectively integrate knowledge from three sources: the model's parametric memory, external structured knowledge, and external unstructured knowledge.
Most existing prompting methods either rely on one or two of these sources, or require repeatedly invoking large language models to generate similar or identical content.
In this work, we overcome these limitations by introducing a novel semi-structured prompting approach that seamlessly integrates the model's parametric memory with unstructured knowledge from text documents and structured knowledge from knowledge graphs. 
Experimental results on open-domain multi-hop question answering datasets demonstrate that our prompting method significantly surpasses existing techniques, even exceeding those that require fine-tuning.
\end{abstract}
\section{Introduction}
\label{sec:intro}

Large language models (LLMs) have demonstrated an impressive breadth of capabilities in the field of natural language processing (NLP). 
LLMs can be adapted to achieve strong performance on a wide variety of tasks without additional training by \textit{few-shot prompting}: conditioning generation on instructions and several exemplars \citep{brown2020language}.
However, few-shot prompting may produce hallucinations due to an under-representation of knowledge in training datasets \citep{openai2023gpt4}, which compromises their suitability for tasks in which a high degree of factual accuracy is necessary.

Recent works have explored alternative prompting strategies to mitigate this issue in knowledge-intensive tasks. 
These include guiding the model to generate intermediate steps before producing the final answer  \cite{wei2022chain,wang2023selfconsistency} and using external tools, such as information retrievers, to utilize external knowledge \cite{yao2023react,trivedi-etal-2023-interleaving}. 
These strategies have somewhat addressed LLM hallucination issues and have enhanced the model's capability to tackle knowledge-intensive tasks. 
However, there are still limitations to these prompting techniques.

First, current methods do not fully utilize all available knowledge sources. 
Knowledge sources can be categorized into parametric memory inside the model, and structured or unstructured external knowledge. 
Parametric memory is learned from large datasets during pre-training and stored in the model's parameters. 
Unstructured external knowledge typically refers to text-based knowledge, such as paragraphs from Wikipedia documents, while structured external knowledge is usually in the form of tables or knowledge graphs such as Wikidata \cite{vrandevcic2014wikidata}. 
Most existing prompting strategies utilize only one or two types of knowledge sources. For example, Chain-of-Thought \cite[CoT;][]{wei2022chain} and several follow-up works \cite{zhang2023automatic,yao2023tree,long2023large} mainly focus on invoking knowledge stored internally in the model's parametric memory, which cannot be updated without further training. 
Other methods like \citet{shao2023enhancing} and \citet{trivedi-etal-2023-interleaving} attempt to integrate parametric memory within the model with external unstructured text, by making multiple calls to LLMs and using external information retrievers to search for documents relevant to the task.
These approaches omit the large amount of information stored in knowledge graphs.

Second, there is a lack of seamless synergy between LLMs and external tools. 
This leads to costly repeated calls to LLMs and post-processing of external tool results during the reasoning process. 
Consequently, most recent methods have only been tested on a limited number of examples in small-scale experiments, such as in \citet{trivedi-etal-2023-interleaving}, \citet{yoran2023answering} and \citet{jiang2023active}.
For instance, \citet{gou2023critic}, \citet{yao2023react}, and \citet{wang2023boosting} use external tools to provide feedback on content generated by LLMs, and then repeatedly invoke LLMs to regenerate content based on this feedback. 
This often results in the model regenerating a large amount of the same content, which is inefficient and difficult to scale. 
Moreover, new hallucinations may be introduced in the regeneration process.

Our work addresses the research question of how to efficiently integrate the three primary knowledge sources during inference time: parametric memory of LLMs, external structured knowledge, and external unstructured knowledge.
We propose a semi-structured chain of thought method, focusing on multi-hop reasoning question answering. 
First, we use an LLM with few-shot prompting to parse the question into a masked semi-structured reasoning chain, followed by syntactic-based filtering to retain only those reasoning chains that are syntactically correct.
Then, we use external tools such as document retrievers and entity linkers to query external knowledge sources to sequentially fill in the masks within the reasoning chain.
Finally, we call upon LLMs as needed to fill in any remaining masks to arrive at the final answer. 
Our approach obviates the need for the LLM to repeatedly generate a large amount of redundant content and seamlessly synergizes all knowledge sources through the semi-structured reasoning chain to answer the question.

To demonstrate the effectiveness of our approach, we conduct extensive experiments using open-source LLMs across various model sizes on several multi-hop question answering datasets. 
Our contributions are as follows:
\begin{enumerate}
    \item We propose a simple, intuitive, and efficient inference-time method to integrate various sources of knowledge for reasoning.
    \item We compare our method with existing approaches on multi-hop question answering datasets. Our method achieves state-of-the-art performance, surpassing even those alternatives that require supervised fine-tuning.
    \item We conduct detailed analyses to investigate the significance of each element of our method and make our code publicly available.\footnote{\url{https://github.com/IntelLabs/multimodal_cognitive_ai/tree/main/Semi-Structured-CoT}}
\end{enumerate}
\section{Related work}
\label{sec:related}

\paragraph{LLM reasoning with CoT.}
Since \citet{wei2022chain} first proposed CoT prompting, a variety of CoT-style approaches have been proposed to further improve the reasoning capabilities of LLMs. Whereas the original CoT utilized greedy decoding, \citet{wang2023selfconsistency} showed that prediction accuracy can be improved by sampling a diverse set of CoT reasoning paths and selecting the final answer from the plurality of predictions among samples. \citet{kojima2022large} demonstrated that even zero-shot CoT significantly improves the reasoning abilities of LLMs across a variety of tasks. \citet{madaan2022text} proposed a concise CoT that prunes intermediate steps to only contain key text and patterns necessary for reasoning. 

Least-to-most prompting \citep{zhou2023leasttomost} extends the ability of CoT to generalize to harder problems than seen in examples by breaking down complex reasoning tasks into easier sub-problems. \citet{yoran2023answering} propose multi-chain reasoning, which reasons over multiple CoTs rather than only aggregating their answers. Tree of Thoughts \citep{yao2023tree} also considers multiple reasoning paths while enabling looking ahead and backtracking through self-evaluation. In contrast with our Semi-CoT methodology, the aforementioned approaches rely solely on the use of the LLMs parametric memory for reasoning and do not integrate CoT reasoning with external knowledge sources.

\paragraph{Retrieval-augmented approaches for multi-hop reasoning.}
To address the lack of knowledge necessary for complex reasoning tasks in the parametric memory of LLMs, several methods that retrieve information from external sources have been proposed. \citet{press2023measuring} introduced Self-Ask, which prompts the LLM to ask follow-up questions which can be answered by external search engines. Demonstrate-Search-Predict (DSP) \citep{Khattab2022DemonstrateSearchPredictCR} breaks down multi-hop questions into sub-problems, leveraging retrieval models to help generate intermediate answers using external documents. ReAct \citep{yao2022react} and IRCoT \citep{trivedi-etal-2023-interleaving} interleave CoT with the retrieval of documents from external sources to reduce hallucinations and improve accuracy on multi-hop QA. Iter-RetGen \citep{shao2023enhancing} uses complete model outputs to retrieve relevant documents over multiple iterations to refine the answer. 

\begin{figure*}
    \centering
    \includegraphics[scale=0.51]{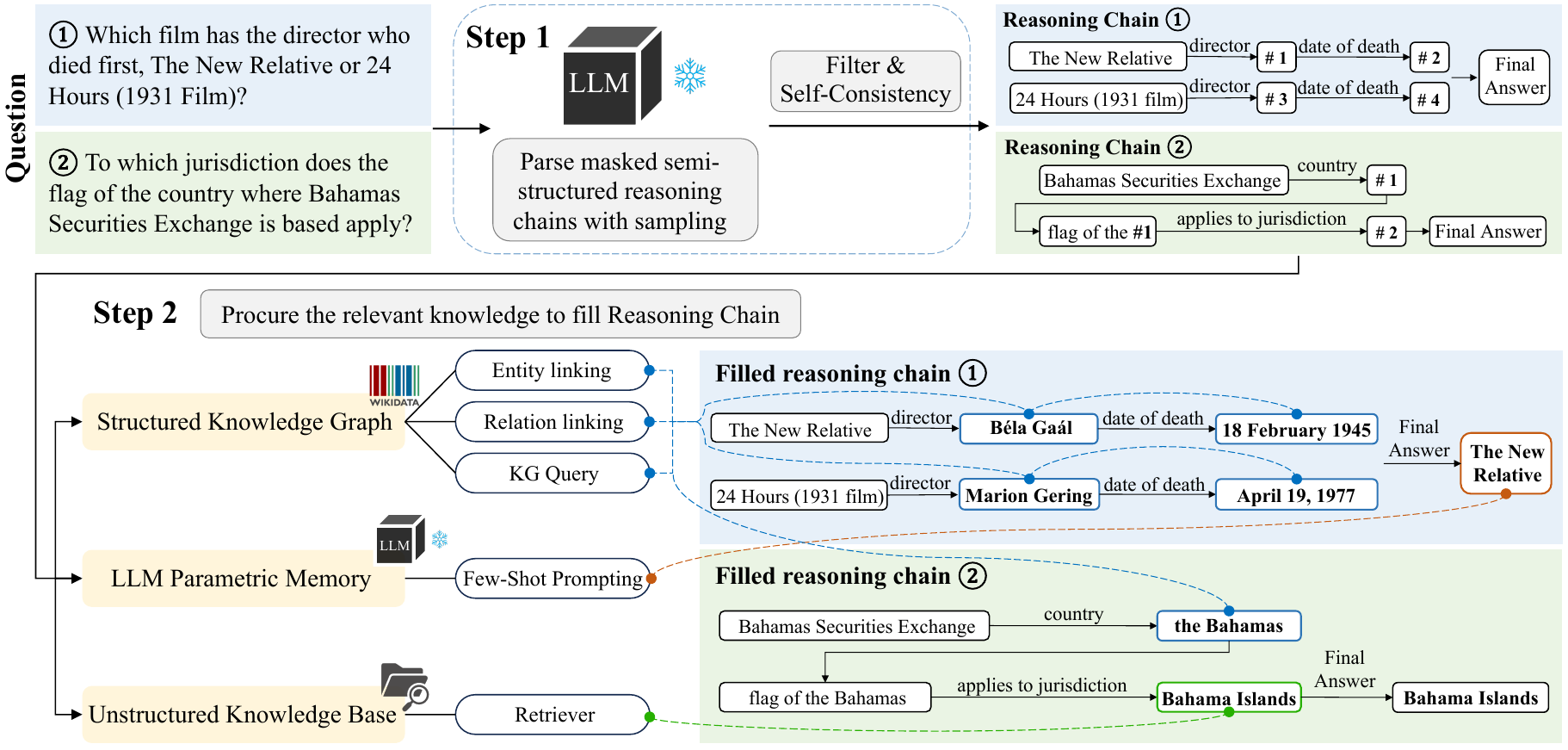}
    \caption{Overview of our approach using different sources of knowledge.}
    \label{fig:method-overview}
\end{figure*}

Verify-and-edit \citep{zhao-etal-2023-verify} identifies cases where sampled CoTs have lower-than-average consistency and then retrieves external knowledge from search engines and document repositories to edit the rationales generated by the LLM. \citet{jiang2023active} also propose an adaptive approach for retrieving external knowledge by identifying the presence of low-probability tokens in intermediate steps. LLM-Augmenter \citep{peng2023check} uses multiple sub-modules to iteratively refine generations through fact checking and retrieval of supporting information from external knowledge sources. These approaches differ from ours primarily in their exclusive focus on integrating external knowledge from text sources, whereas our use of a semi-structured CoT format enables the seamless integration of both unstructured text and structured knowledge sources.

\paragraph{Structured knowledge integration with LLMs.}
Relatively little prior work has explored strategies for integrating structured knowledge with frozen LLMs at inference time. GenRL \cite{rossiello2021generative} integrates structured data into the model's input to train a generative relation linking model. ERNIE 3.0 \citep{sun2021ernie} and SKILL \citep{moiseev-etal-2022-skill} both train LLMs using triples from structured knowledge graphs to infuse knowledge into the LLM's parametric memory. QA-GNN \citep{yasunaga-etal-2021-qa} and GreaseLM \citep{zhang2022greaselm} integrate knowledge graph embeddings with language model encodings through the use of GNNs. Recently, \citet{jiang2023structgpt} proposed an iterative approach for retrieving and reasoning over information from structured knowledge sources during LLM inference by linearizing retrieved knowledge into text. In contrast, our approach facilitates the integration of both unstructured text and structured knowledge without requiring specialized linearization interfaces.

\section{Methodology}
\label{sec:method}

\paragraph{Overview}
We focus on multi-hop question answering tasks. To synthesize the answers to a given multi-hop question, we synergistically integrate an LLM, structured knowledge from an external Knowledge Graph (KG), and unstructured textual knowledge from an external text knowledge base (TKB).
Our method unfolds in two steps:
\begin{enumerate}
    \item We prompt an LLM to parse the input question into a semi-structured reasoning chain with masks as placeholders. 
    \item We procure knowledge from three sources to fill in the masks in the reasoning chain, thereby deriving the final answer. We leverage knowledge from the LLM via few-shot prompting, use entity and relation linking models to ground the reasoning chain to the KG, and employ a dense retriever to fetch relevant documents from the TKB.
\end{enumerate}
No components of our method require any additional training. A major strength of our method is its simplicity, intuitiveness, and ease of use. 
We present an overview of our method in \Cref{fig:method-overview}.

\subsection{Semi-Structured Reasoning Chain Parsing}
Translating a question into a semi-structured reasoning chain can be seen as a semantic parsing task, where the multi-hop question serves as the utterance, and the reasoning chain serves as the logical form. 
Unlike standard semantic parsing tasks such as text2SQL, the logical forms we parse cannot be directly used to retrieve answers from a database. 
Instead, the subsequent steps involve populating the masked placeholders in the reasoning chains with accurate knowledge to derive the final answer.

Drawing inspiration from \citet{brassard-etal-2022-copa} and the annotation of \citet{trivedi-etal-2022-musique}, we posit that a good reasoning chain should use the most relevant set of facts to coherently connect the question to the answer, creating a minimal bridge between the two. 
To establish this bridge, our reasoning chain uses a series of triplet structures, each expressed as \verb|(head, relation, tail)|.
The \verb|head|, \verb|relation|, and \verb|tail| can appear as specific entity mentions or relations in a knowledge graph, as free-form text, or as masks symbolized by a \verb|#| followed by an ascending number (e.g., \verb|#1|, \verb|#2|) indicating parts that need to be populated. 
These masks can be either of the two aforementioned forms. 
The reasoning chain ends with a masked final answer, denoted as \verb|final answer: #answer| or \verb|final answer: #<number>| based on specific reasoning logic.
This blended structure allows us to merge structured and unstructured knowledge, creating a unified reasoning pathway.

We use an LLM to parse masked reasoning chains from each input question with few-shot prompting.
For example, for the question:
\begin{quote}
    "To which jurisdiction does the flag of the country where the Bahamas Securities Exchange is based apply?",
\end{quote}
the corresponding reasoning chain is: 
\begin{quote}
    "(Bahamas Securities Exchange, country, \verb|#1|); (flag of the \verb|#1|, applies to jurisdiction, \verb|#2|); final answer: \verb|#2|"
\end{quote}
A natural question is: why not adhere strictly to either structured or unstructured formats? The answer lies in the varying complexity of grounding different knowledge pieces to their sources, and the extent to which these sources cover the knowledge. 

For example, understanding the positions held by George Washington through his Wikipedia page requires a deep dive into a long document, semantic understanding, and temporal reasoning. 
While challenging to associate with unstructured sources, this knowledge is easily depicted in the Wikidata knowledge graph with a series of triples, such as (George Washington, position held, Commander-in-Chief) (as shown in the example in \Cref{sec:appendix-struct-vs-unstruct} \Cref{fig:struct-vs-unstruct}). 
Nevertheless, the scope of knowledge graphs is finite; we cannot encapsulate every fact in a structured form. Therefore, semi-structured triplets with free-form text are also essential to ground knowledge with other unstructured sources.

As in other semantic parsing tasks, LLMs can produce both syntactic and semantic errors when parsing reasoning chains.
We define those reasoning chains that do not follow the pre-defined format as reasoning chains with "syntactic errors".
We define reasoning chains that are syntactically correct but logically flawed, meaning the reasoning chains cannot coherently connect the question to the answer, as "semantic errors".
To address syntactic errors, we employ the LLM to parse multiple reasoning chains for each input question through sampling. We then filter out any chains that have syntactic errors (see \Cref{sec:appendix-syntactic-errors} for the types of errors we filter). Finally, we apply the Self-Consistency strategy \cite{wang2023selfconsistency} to derive the final masked semi-structured reasoning chain.
We leave addressing semantic errors to future work.

\subsection{Masked Reasoning Chain Filling}
We procure factual knowledge from three sources to fill the masks within parsed reasoning chains: the LLM's parametric memory, the structured triples of the KG, and the unstructured documents from the TKB. We generally prioritize parametric memory last due to its lower reliability in knowledge retrieval. The choice between structured and unstructured knowledge often depends on the predominant knowledge type required for the specific task. In our experiments, we have explored all possible sequences of their application. 
We iteratively fill the masks in each reasoning chain triplet from left to right, leveraging these knowledge sources. If we encounter instances where the triplets in the chains cannot be grounded to the available structured or unstructured knowledge, we leave those masks unfilled. In each iteration, we extract valid triplets (where both the head and the relation are unmasked, and only the tail is masked) from the generated masked reasoning chains for filling.

\paragraph{Use of Structured Knowledge}
\Cref{alg:kg-fill} presents our algorithm for filling the reasoning chain using a structured knowledge source. For the KG grounding of a specific triplet (line \ref{kg-ground}), we take a triplet from \Cref{fig:method-overview} as an example: (24 Hours (1931 film), director, \verb|#3|). We begin by conducting entity linking for "24 Hours (1931 film)" and relation linking for "director." We then query the KG to identify the entity corresponding to "\verb|#3|", which is Marion Gering. This allows us to fill in the reasoning chain's "\verb|#3|" mask. 

\begin{algorithm}[t]
\SetAlgoLined
\small
\KwIn{\\
\quad\(RC\): the generated masked reasoning chains\;
\quad\(KG\): the external knowledge graph\;
\quad\(EL\): the entity linking model\;
\quad\(RL\): the relation linking model\;}

\(T_{q} \gets \emptyset\)\;

\While{\textbf{true}}{
    \(T_v \gets  \text{extract valid triplets from } RC\)\;
    \(T_v \gets T_v \setminus \{t \mid t \in T_{q} \}\)\; 
    \If{\(T_v = \emptyset\)}{
        \textbf{break}; 
    }
    \(T_q \gets T_q \cup T_v \)\;
    \(T_g \gets \) ground each triplet in \(T_v\) onto \(KG\) using \(EL\) and \(RL\)\; \label{kg-ground}

    \If{\(T_g \neq \emptyset\)}{
        \(T_a \gets\) Query \(KG\) using \(T_g\)\;
        \If{\(T_a \neq \emptyset\)}{
            \(RC \gets \) fill \(RC\) with information from \(T_a\)\;
        }
    }
}

\Return{\(RC\)}

\caption{Fill the Masked Reasoning Chains with Structured Knowledge}
\label{alg:kg-fill}
\end{algorithm}

\paragraph{Use of Unstructured Knowledge}
\Cref{alg:text-fill}  presents our algorithm for filling reasoning chains using unstructured textual knowledge. 
Here, our approach for filling each valid triplet adheres to the classic retrieve-and-read strategy. For instance, consider the triplet from \Cref{fig:method-overview}: (Bahamas Securities Exchange, country, \verb|#1|). We first prompt the LLM to convert it into a straightforward single-hop question: In which country is the Bahamas Securities Exchange located? Next, we employ the retriever to fetch relevant documents from the TKB. Finally, we combine the retrieved documents with the question and prompt the LLM again to obtain the answer: the Bahamas.

\begin{algorithm}[t]
\SetAlgoLined
\small
\KwIn{\\
\quad\(LLM\): the pre-trained language model\;
\quad\(RC\): the generated masked reasoning chains\;
\quad\(TKB\): the external unstructured text knowledge base\;
\quad\(Retriever\): the dense retriever\;
\quad\(k\): the number of top documents to retrieve\;
}

\(T_{q} \gets \emptyset\)\;

\While{\textbf{true}}{
    \(T_v \gets  \text{Extract valid triplets from } RC\)\;
    \(T_v \gets T_v \setminus \{t \mid t \in T_{q} \}\)\; 
    \If{\(T_v = \emptyset\)}{
        \textbf{break}; 
    }
    \(T_q \gets T_q \cup T_v \)\;
    \(Q_{gen} \gets \) few-shot prompt \(LLM\) to generate single-hop questions for each triplet in \(T_v\)\;
    \(D_{ret} \gets \) use \(Retriever\) to retrieve the top \(k\) documents for each question in \(Q_{gen}\) from \(TKB\)\;
    \(T_a \gets \) few-shot prompt \(LLM\) to answer each question in \(Q_{gen} \) based on corresponding documents in \(D_{ret}\)\;
    \If{\(T_a \neq \emptyset\)}{
        \(RC \gets \) fill \(RC\) with information from \(T_a\)\;
    }
}

\Return{\(RC\)}

\caption{Fill the Masked Reasoning Chains with Unstructured Knowledge}
\label{alg:text-fill}
\end{algorithm}

\paragraph{Use of Parametric Memory}
We utilize the parametric memory of the LLM in the final step of our method. Specifically, we employ few-shot prompting to have the LLM fill in any remaining masks. For instance, in the case of \Cref{fig:method-overview}, question 1, even after all the masks in the reasoning chain have been filled, we may still need to rely on the LLM for a final comparison of time to arrive at the final answer.
\section{Experiments}
\label{sec:experiments}

\subsection{Datasets}
We evaluate our proposed method on three knowledge-intensive multi-hop reasoning datasets: 2WikiMultihopQA \cite[2Wiki;][]{ho-etal-2020-constructing},  MuSiQue-Ans \cite[MuSiQue;][]{trivedi-etal-2022-musique} and Bamboogle \cite{press2023measuring}.
Following previous works \cite{press2023measuring,shao2023enhancing,chen2023efficient}, due to the unavailability of labels for the test sets of 2Wiki and MuSiQue, we utilize their training sets to develop our prompts and the development sets as our test sets. Bamboogle only provides a test set, and we test our method on its entire test set. Additional dataset details are in \Cref{sec:appendix-datasets}.

\paragraph{Evaluation Metrics}
For the 2Wiki and MuSiQue datasets, we employ the official evaluation methods from their respective code bases to compute the Exact Match (EM) and F1 scores for the answers. The Bamboogle code base does not offer an official evaluation method. Instead, we use the evaluation code from MuSiQue to calculate the metrics.

\subsection{Implementation Details}

\paragraph{Base Models}
In all of our experiments, we use few-shot in-context learning and do not perform any model training. 
We utilize three different sizes of language models from the LLAMA 2 family (7b, 13b, and 70b)  \cite{touvron2023LLAMA} as our base models because of their strong empirical results and open-source availability. Additionally, we also use the LLAMA-65b model in some of our experiments for a fair comparison with previous methods.
We employ these base models to parse semi-structured reasoning chains, convert masked triplets into single-hop questions for answering, and fill in reasoning chains.

\paragraph{Prompts}
All of our in-context learning exemplars are randomly sampled from the training set of the corresponding dataset, and their semi-structured reasoning chains are manually annotated by the authors. Since Bamboogle does not have a training set, we use examples from the MuSiQue training set for this purpose. More details about the prompts can be found in \Cref{sec:appendix-ic-examples}.

\paragraph{Knowledge Sources and Usage}
2Wiki and MuSiQue were initially created for reading comprehension, with each question accompanied by multiple context paragraphs, including supporting and distracting paragraphs. Following \citet{chen2023efficient} and \citet{trivedi-etal-2023-interleaving}, we adapt them for an open-domain setting by collecting all context paragraphs from the training, development, and test sets' questions to serve as an unstructured TKB.
Bamboogle was originally an open-domain setup. We follow the original setup and use Google Search API \footnote{\url{https://serpapi.com/}} to access the entire web as the unstructured TKB. We host a local Wikidata \footnote{We use a snapshot of Wikidata from January 20, 2023.} \cite{vrandevcic2014wikidata} endpoint as our structured knowledge source for all the datasets. 
We employed the pre-trained Contriever (\verb|Contriever-msmarco|) \cite{izacard2022unsupervised} for document retrieval, the BLINK model \cite{wu-etal-2020-scalable} for entity linking, and the all-MiniLM-L6-v2 model from SentenceTransformers \cite{reimers-gurevych-2019-sentence} for relation linking. 
See \Cref{sec:appendix-knowledge-source-and-useage-details} for more details. 

\begin{table*}[t]
    \small
    \centering
    \setlength{\tabcolsep}{0.4em}
    \resizebox{0.80\textwidth}{!}{
    \begin{tabular}{l c l c c c c}
        \toprule
        & & & \multicolumn{2}{c}{\textbf{2Wiki}} & \multicolumn{2}{c}{\textbf{MuSiQue}}  \\
        \cmidrule(lr){4-5}
        \cmidrule(lr){6-7}
        \textbf{Method} & \textbf{Inference Only} & \textbf{Base Model} & \textbf{EM} & \textbf{F1} & \textbf{EM} & \textbf{F1} \\
        \midrule
        Self-Ask $^{*}$ & \Checkmark & text-davinci-003 & 0.37 & 0.49 & 0.28 & 0.42 \\
        CoT $^{*}$ & \Checkmark & text-davinci-003 & 0.30 & 0.40 & 0.19 & 0.31 \\
        ITER-RETGEN $^{*}$ & \Checkmark& text-davinci-003 & 0.36 & 0.47 & 0.26 & 0.42 \\
        ReAct $^{*}$  & \Checkmark & text-davinci-003 & 0.28 & 0.39 & 0.23 & 0.37 \\
        \midrule
        Data Synthesis + SC $^{\diamondsuit}$ & \XSolidBrush & llama-65b & 0.51 & 0.60 & 0.31 & 0.42 \\
        DSP $^{\diamondsuit}$ & \Checkmark & llama-65b & 0.36 & 0.44 & 0.21 & 0.29 \\
        Self-Ask $^{\diamondsuit}$ & \Checkmark & llama-65b & 0.35 & 0.42 & 0.20 & 0.28 \\
        \midrule
        \multicolumn{7}{c}{\emph{Ours}} \\
        \midrule
        Standard & \Checkmark & llama-65b   & 0.29 & 0.34 & 0.11 & 0.21 \\
        Text + KG + Model & \Checkmark & llama-65b   & 0.58 & 0.63 & 0.32 & 0.41 \\
        KG + Text + Model & \Checkmark & llama-65b   & 0.78 & 0.82 & 0.32 & 0.40 \\
        \midrule
        Standard & \Checkmark & llama2-7b & 0.27 & 0.33 & 0.07 & 0.15 \\
        Text + KG + Model & \Checkmark & llama2-7b & 0.44 & 0.49 & 0.24 & 0.33 \\
        KG + Text + Model & \Checkmark & llama2-7b & 0.65 & 0.70 & 0.25 & 0.34 \\
        \midrule
        Standard & \Checkmark & llama2-13b & 0.28 & 0.34 & 0.09 & 0.18  \\
        Text + KG + Model & \Checkmark & llama2-13b & 0.50 & 0.56 & 0.32 & 0.40 \\
        KG + Text + Model & \Checkmark & llama2-13b & 0.68 & 0.73 & 0.31 & 0.40 \\
        \midrule
        Standard & \Checkmark & llama2-70b & 0.33 & 0.39 & 0.14 & 0.24 \\
        Text + KG + Model & \Checkmark & llama2-70b & 0.73 & 0.77 & \textbf{0.40} & \textbf{0.50} \\
        KG + Text + Model & \Checkmark & llama2-70b & \textbf{0.82} & \textbf{0.87} & 0.39 & 0.48 \\
        \bottomrule
    \end{tabular}
    }
    \vspace{1mm}
    \caption{Evaluation results on 2Wiki and MuSiQue datasets. The highest performance is \textbf{bolded}. SC is self-consistency \cite{wang2023selfconsistency}. Text/KG/Model refers to the use of retrieved paragraphs/the knowledge graph/model's parametric knowledge to fill the reasoning chains. 
    $*$ indicates results from \citet{shao2023enhancing}, who used a smaller-scale evaluation. We use their results for comparison to identify the strongest baselines.
    $\diamondsuit$ indicates results from \citet{chen2023efficient}, who employed the same evaluation settings as ours.}
    \label{tab:main-results}
\end{table*}

\begin{table}[t]
    \small
    \centering
    \setlength{\tabcolsep}{0.4em}
    \begin{tabular}{l c c}
        \toprule
        \textbf{Method} & \textbf{EM} & \textbf{F1}\\
        \midrule
        Self-Ask $^{\heartsuit}$ & 0.49 & 0.62 \\
        \midrule
        Standard &  0.23 & 0.34\\
        Text + KG + Model &  \textbf{0.54} & 0.67 \\
        KG + Text + Model &  \textbf{0.54} & \textbf{0.69} \\
        \bottomrule
    \end{tabular}
    \vspace{1mm}
    \caption{Evaluation results on Bamboogle dataset using LLAMA2-70b as the base model. The highest performance is \textbf{bolded}. Text/KG/Model refers to the use of retrieved paragraphs/the knowledge graph/model's parametric knowledge to fill the reasoning chains. $\heartsuit$ indicates our replication.}
    \label{tab:bamboogle-results}
\end{table}

\subsection{Baselines}
On 2Wiki and MuSiQuem datasets, we compare our proposed method with the following state-of-the-art prompting methods: standard few-shot prompting (standard) \cite{NEURIPS2020_1457c0d6}, CoT prompting \cite{wei2022chain}, Self-Ask \cite{press2023measuring}, ITER-RETGEN \cite{shao2023enhancing}, ReAct \cite{yao2023react}, and DSP \cite{Khattab2022DemonstrateSearchPredictCR}. We also contrast our method with Data Synthesis \cite{chen2023efficient}, which requires fine-tuning. Data Synthesis generates millions of synthetic question-answer pairs and fine-tunes the base LLM on the generated data, utilizing a prompting method similar to ReAct. More detailed baseline descriptions are provided in \Cref{sec:appendix-baselines-details}.

The Bamboogle dataset does not provide an official evaluation method\footnote{The original work relies on human evaluation for model performance.}, which makes it challenging to directly compare our results with previously reported model performance. Considering the cost of Google Search API call, we replicate the best-performing baselines, Self-Ask (see analysis in \Cref{baseline-comparsion}), based on the LLAMA2-70b model on Bamboogle for a fair comparison\footnote{Our replication of Self-Ask uses the officially released code: \url{https://github.com/ofirpress/self-ask}.}.

\subsection{Main Results}
\label{section:main-results}

We present the performance comparison between the baselines and our method in \Cref{tab:main-results,tab:bamboogle-results}.

\paragraph{Which baseline methods perform best?}
\label{baseline-comparsion}
We compare the baselines on larger datasets, 2Wiki and MuSiQue, as shown in the first two sections of \Cref{tab:main-results}. 
For those methods not involving fine-tuning, Self-Ask is particularly effective. 
When using the text-davinci-003 as the base model, Self-Ask excels in all metrics across all datasets, indicating that simply generating clear multi-round sub-questions and retrieving relevant documents is more effective than more complex methods such as ITER-RETGEN. 
This holds true when Self-Ask is applied to the LLAMA-65b model, where it also performs on par with DSP.
On the other hand, the Data Synthesis method, fine-tuned on more than one million synthesized question-answer pairs, surpasses other inference-only methods in all metrics on both datasets. The weakest performance is observed in the standard few-shot prompting method, which does not generate any intermediate steps.

\paragraph{Does our method outperform the baseline methods?}
Our method significantly improves over the standard few-shot prompting across all models, datasets, and metrics. For example, compared to the standard few-shot prompting, our KG+Text+Model method improves the exact match score by 148\%, 179\%, and 135\% on the 2Wiki, MuSiQue, and Bamboogle datasets, respectively. 

When compared with the existing state-of-the-art inference-only prompting method, Self-Ask, our method still exhibits substantial superiority across all models, datasets, and metrics. When using the same base model, our KG+Text+Model method surpasses the exact match score of Self-Ask by 120\% and 60\% on the 2Wiki and MuSiQue datasets, respectively (based on LLAMA-65B), and also exceeds it by 10\% on the Bamboogle dataset (based on LLAMA-70B).

Compared to the Data Synthesis+SC method that involves fine-tuning, our KG+Text+Model approach surpasses it by 53\% in exact match score on 2Wiki, using the same base LLM, LLAMA-65b. Additionally, our method shows very similar performance on MuSiQue, with a slightly higher exact match and a slightly lower F1 score. When employing LLAMA2-70b as the base model, our approach again significantly outperforms the Data Synthesis+SC method on both datasets.

\paragraph{Does the order of using knowledge sources matter?}
The impact of using knowledge sources in different sequences varies across datasets. 
For the 2Wiki dataset, prioritizing structured knowledge from knowledge graphs to fill reasoning chains proves to be significantly more effective than starting with retrieved textual knowledge, regardless of the model size. 
This can likely be attributed to the dataset's integration of a substantial amount of structured knowledge at the time of its creation, making its questions more amenable to decomposition into structured triplets.
In this case, entity linking models can more easily identify and match structured knowledge.
Hence, for 2Wiki, knowledge graphs may serve as a more reliable source of knowledge compared to retrieved text knowledge. 
However, the MuSiQue and Bamboogle datasets, which mainly consist of a composition of single-hop questions, seem less sensitive to the preference for knowledge sources, rendering the sequence in which knowledge is used less impactful.

\paragraph{Does the model size matter?}
Our method benefits from increased model size, resulting in continuous performance improvements, regardless of the order in which sources of knowledge are utilized. 
This trend aligns with the majority of current LLM prompting methods. 
Notably, our method also significantly boosts the performance of the smaller LLAMA2-7b model, allowing it to surpass the LLAMA2-70b model with standard few-shot prompting, which is ten times larger, across all datasets and metrics.
\section{Analysis}
\label{sec:analysis}

\paragraph{What is the effect of different knowledge sources?}
 We conduct ablation experiments on all the datasets, with the results presented in \Cref{table:knowledge-souces-comp}. 
 As discussed in \Cref{section:main-results}, the characteristics of the dataset significantly influence the contributions of the knowledge sources to the final performance of the method.
On the 2Wiki dataset, using the KG and LLM's parametric memory yields comparable results to using all three knowledge sources. However, for MuSiQue and Bamboogle datasets, combining external text knowledge with model memory alone was enough to match or exceed the performance of using all knowledge sources.

\begin{table}[t]
\small
\centering
\setlength{\tabcolsep}{0.4em}
\begin{tabular}{l c c c c c c}
\toprule
&  \multicolumn{2}{c}{\textbf{2Wiki}} & \multicolumn{2}{c}{\textbf{Mu}} & \multicolumn{2}{c}{\textbf{Ba}} \\
\cmidrule(lr){2-3}
\cmidrule(lr){4-5}
\cmidrule(lr){6-7}
Model & EM & F1 & EM & F1 & EM & F1\\
\midrule
Model & 0.46 & 0.52 & 0.24 & 0.33 & 0.42 & 0.55 \\
Text + Model & 0.62 & 0.67 & 0.39 & 0.49 & \textbf{0.57} & \textbf{0.70} \\
KG + Model & 0.81 & 0.85 & 0.28 & 0.38 & 0.40 & 0.55 \\
KG + Text + Model &  \textbf{0.82} & \textbf{0.87} & 0.39 & 0.48 & 0.54 & 0.69 \\
Text + KG + Model & 0.73 & 0.77 & \textbf{0.40} & \textbf{0.50} & 0.54 & 0.67 \\
\bottomrule
\end{tabular}
\caption{Performance of using different knowledge sources on 2Wiki, MuSiQue (Mu), and Bamboogle (Ba) with LLAMA2-70b as the base model.}
\label{table:knowledge-souces-comp}
\end{table}

\paragraph{What is the impact of using an oracle KG?}
The 2Wiki dataset provides relation triples as gold evidence for each question. We utilize these triples to construct an Oracle KG, significantly smaller than Wikidata, and employ string matching to link the triplets in our reasoning chain to the Oracle KG. This approach aims to simulate our method's performance with enhanced entity linking and relation linking. The results are presented in \Cref{sec:appendix-oracle-kg} \Cref{table:oracle-kg-comp}. Using the Oracle KG indeed improves performance, but the extent of this improvement is considerably smaller compared to the enhancement our method achieves over standard few-shot prompting techniques and other state-of-the-art prompting methods. This outcome also suggests that our approach can effectively leverage large-scale KGs without the need for task-specific fine-tuning of the entity linking model.

\paragraph{What are the errors in our method?}
To better understand the errors of our proposed method, we run our LLAMA2-70b-based KG+Text+Model method on a sample of 200 training instances from the 2Wiki and MuSiQue datasets. We manually annotate 37 and 50 errors sampled from these instances (totaling 87 errors). (In 2Wiki, there are only 37 errors.) \Cref{tab:error-analysis} in \Cref{sec:appendix-error-analysis} provides the error types and corresponding examples. In the 2Wiki dataset, the majority of the errors, approximately 78\%, are false negatives, where the model prediction are alternative phrasings of the gold answers (e.g., the model predicts ``1'', but the reference answer is ``one''). The remaining errors include 19\% knowledge retrieval errors, where the model fails to correctly utilize relevant knowledge to fill the reasoning chain (e.g., failing to retrieve the correct dates), and 3\% reasoning chain errors, where the model's initial reasoning chain is incorrect (e.g., confusing who in the question a requested position refers to). However, in the MuSiQue dataset, 52\% of errors are due to knowledge retrieval, 26\% are reasoning chain errors, and 22\% are false negatives. These analyses suggest that our method might be underestimated for the 2Wiki dataset, which relies more on structured knowledge. For datasets like MuSiQue, which depend more on unstructured knowledge, performance could be significantly improved by fine-tuning knowledge retrieval methods, such as document retrievers, and filtering out reasoning chains with semantic errors.
\section{Conclusion}
\label{sec:conclusion}

In this work, we explored integrating knowledge from three different sources through semi-structured reasoning chains. We found that our approach significantly improved the performance of LLMs on multi-hop reasoning tasks, surpassing other state-of-the-art prompting methods. In future work, we plan to improve our method through semantic error filtering in reasoning chain parsing and knowledge retriever fine-tuning. Moreover, we aim to investigate more intelligent knowledge source selection methods to mitigate potential conflicts among the knowledge sources.

\section*{Limitations}
\label{sec:limitations}
In this work, we exclusively investigate utilizing large language models from the open-source LLAMA family as our base models. We do not examine other potentially more robust open-source or proprietary models. We also only focus on the inference capabilities of these models without engaging in any training or fine-tuning processes. It is likely that training could improve performance, though it would require significant additional computational resources.
\section*{Ethics Statement}
Wikipedia contains certain biases \cite{falenska-cetinoglu-2021-assessing}, and we use data from Wikipedia in the text knowledge bases and knowledge graphs for retrieval, so we are potentially introducing similar biases into our method.

\bibliography{anthology,custom}
\clearpage
\section*{Appendix}
\label{sec:appendix}
\appendix

\section{Comparison of Structured and Unstructured Knowledge Sources}
\label{sec:appendix-struct-vs-unstruct}
\begin{figure}[h]
    \centering
    \includegraphics[scale=0.70]{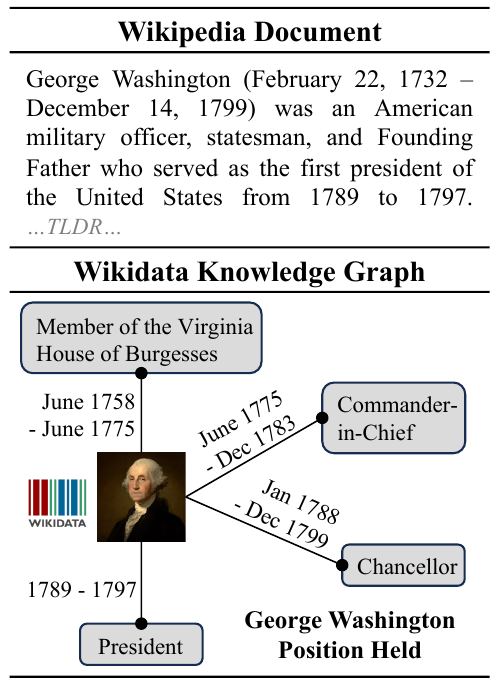}
    \caption{Comparison of structured and unstructured knowledge.}
    \label{fig:struct-vs-unstruct}
\end{figure}

In \Cref{fig:struct-vs-unstruct}, we show that sometimes structured knowledge is easier to use compared to unstructured knowledge.

\section{Pre-defined Syntactic Errors}

\label{sec:appendix-syntactic-errors}
When filtering sampled masked reasoning chains, we filter out chains that contain the following syntactic errors.

\begin{itemize}
    \item \textit{\textbf{MaskUnderflowError}}/\textit{\textbf{MaskOverflowError}}: The two errors refer to the number of unique masks in the reasoning chain being either below the minimum threshold or above the maximum threshold. By pre-setting the quantity of masks, we can limit the number of masks included in the reasoning chain generated by LLM. The number of hops in a multi-hop question is typically finite. For different tasks, we can set the possible number of masks based on prior knowledge. For instance, in multi-hop question-answering tasks, the reasoning chain should have at least two unique masks and rarely involve more than four hops. Thus, we can set the minimum number of masks to two and the maximum to four, filtering out reasoning chains that don't meet these criteria.
    \item \textit{\textbf{DiscontinuousMaskError}}: This error indicates that the sequence numbers of masks in the reasoning chain are not sequential, such as jumping from \#1 to \#3.
    \item \textit{\textbf{MissingMaskNumberError}}:  This error refers to an issue where, apart from the final answer mask "\#answer", the generated mask lacks a number following the "\#".
    \item \textit{\textbf{MissingTripleMaskError}}: The error refers to the case where neither the head nor the tail in the generated triple is masked.
    \item \textit{\textbf{IncorrectTripleRelationMaskError}}: We define that only the head and tail can be masked. This error points to cases where the relation in a triple is masked.
    \item \textit{\textbf{IncorrectTripleFormatError}}: This error indicates that the generated triple doesn't follow the "head \verb|>>| relation \verb|>>| tail" format. For example, the LLM might split the relation into two parts separated by "\verb|>>|", turning the triple into "head \verb|>>| relation words \verb|>>| relation words \verb|>>| tail."
    \item \textit{\textbf{FinalAnswerFormatError}}: This error is flagged when the generated reasoning chain doesn't conclude with the format "final answer: \#answer."
\end{itemize}

These error definitions can be expanded upon based on specific tasks or results from error analysis. After filtering, we employ the self-consistency method to obtain the final reasoning chain from the filtered chains.

\section{Datasets}
\label{sec:appendix-datasets}
\paragraph{2WikiMultihopQA} 
2WikiMultihopQA is a multi-hop question answering dataset that contains various reasoning types and is developed based on HotpotQA \cite{yang-etal-2018-hotpotqa}. 
However, it excludes the single-hop and context-dependent multi-hop question types. 
The dataset contains 192,606 questions in total, split into 167,454 for training, 12,576 for development, and 12,576 for testing. 
We evaluate our method on the entire development set.

\paragraph{MuSiQue}
MuSiQue-Ans \cite[MuSiQue;][]{trivedi-etal-2022-musique} is comprised of answerable multi-hop questions that are constructed from a large number of single-hop questions in a bottom-up method. 
These questions are designed to force the model to connect all supporting facts for connected reasoning instead of seeking shortcuts to find the answers.
MuSiQue consists of a total of 24,814 questions, split into 19,938 training samples, 2,417 development samples, and 2,459 test samples. 
Following previous works \cite{press2023measuring,shao2023enhancing,chen2023efficient}, we focus on 2-hop questions (1, 252 in the development set).
This is because, as also discussed in \citet{press2023measuring}, many of the automatically composed 3-hop and 4-hop questions in MuSiQue are unnatural and even challenging for humans to comprehend.

\paragraph{Bamboogle}
Bamboogle \cite{press2023measuring} is designed to evaluate the compositionality skills of models. It consists of 125 two-hop questions from various categories, all annotated by humans. These questions are based on Wikipedia but cannot be directly answered through a Google search and have not previously appeared on the web.

\begin{table*}
  \small
  \centering
  \begin{tabular}{p{.95\textwidth}}
    \toprule
    \emph{Parse semi-structured reasoning chains on the MuSiQue Dataset} \\
    \midrule
    Question: What are the spirits associated with Shintoism called in the language of Lala DX? \\
    Reasoning Chain: LaLa DX >> language >> \verb|#|1; spirits associated with Shintoism in \verb|#|1 >> name >> \verb|#|2; final answer: \verb|#|2 \\\\

    Question: When did the person with the famous quote we came we saw we conquered die? \\
    Reasoning Chain: famous quote "we came we saw we conquered" >> author >> \verb|#|1; \verb|#|1 >> date of death >> \verb|#|2; final answer: \verb|#|2 \\
    
    \ldots \\    
    
    Question: Which flag was made first between Cuba and the country with the immigration? \\
    Reasoning Chain: country with the immigration >> name of the country >> \verb|#|1; flag made first between Cuba and \verb|#|1 >> name >> \verb|#|2; final answer: \verb|#|2 \\
    
    \midrule
    \emph{Parse semi-structured reasoning chains on the 2WikiMultiHopQA Dataset}\\
    \midrule
    Question: Were Wessel Dammers and Robert Handcock (Rugby Union) from the same country? \\
    Reasoning Chain: Wessel Dammers >> country of citizenship >> \verb|#|1; Robert Handcock (rugby union) >> country of citizenship >> \verb|#|2; final answer: \verb|#|answer \\\\

    Question: Which award the director of film The Blue Umbrella (2005 Film) won? \\
    Reasoning Chain: The Blue Umbrella >> director >> \verb|#|1; \verb|#|1 >> award received >> \verb|#|2; final answer: \verb|#|answer

    \ldots \\

    Question: Where was the place of death of Strut-Harald's father? \\
    Reasoning Chain: Strut-Harald >> father >> \verb|#|1; \verb|#|1 >> place of death >> \verb|#|2; final answer: \verb|#|answer \\
    
    \midrule
    \emph{Convert triplets to single-hop questions} \\
    \midrule
    
    Triplet: ("basit ali", "place of birth", ?) \\ 
    Question: where was basit ali born? \\\\
    
    Triplet: ("instincts", "performer", ?) \\
    Question: who performed instincts? \\
    
    \ldots \\    
    
    Triplet: ("Notre Dame", "last time won national championship in football", ?) \\
    Question: When was the last time Notre Dame won a national championship in football? \\

    \midrule
    \emph{Answer triplet converted single-hop questions} \\
    \midrule

    Context: The Blue Umbrella (2005 film) \ldots It was directed by Vishal Bhardwaj and starred ldots. \\ 
    Question: Who is the director of The Blue Umbrella? \\
    Answer: Vishal Bhardwaj \\\\

    Context: Marie Louise Coidavid(1778 - 2013 March 11, 1851), was the Queen of the Kingdom of Haiti 1811 \u2013 20 as the spouse of Henri I of Haiti. \\
    Question: Who is the director of The Blue Umbrella? \\
    Answer: unknown \\

    \ldots \\    

    Context: Charles Haughey died of prostate cancer in 2006, at the age of eighty. \\
    Question: What was the cause of death for Charles Haughey? \\
    Answer: prostate cancer \\

    \midrule
    \emph{Use parametric memory to fill the masks} \\
    \midrule

    Question: Were Wessel Dammers and Robert Handcock (Rugby Union) from the same country? \\ Reasoning Chain: Wessel Dammers >> country of citizenship >> \verb|#|1; Robert Handcock (rugby union) >> country of citizenship >> \verb|#|2; final answer: \verb|#|answer \\
    Filled reasoning chain:  Wessel Dammers >> country of citizenship >> Dutch; Robert Handcock (rugby union) >> country of citizenship >> New Zealand; final answer: no \\\\

    Question: Which award the director of film The Blue Umbrella (2005 Film) won? \\
    Reasoning Chain: The Blue Umbrella >> director >> \verb|#|1; \verb|#|1 >> award received >> \verb|#|2; final answer: \verb|#|answer \\
    Filled reasoning chain: The Blue Umbrella >> director >> Vishal Bhardwaj; Vishal Bhardwaj >> award received >> National Film Award for Best Music Direction; final answer: National Film Award for Best Music Direction \\

    \ldots \\  

    Question: Where was the place of death of Strut-Harald's father? \\
    Reasoning Chain: Strut-Harald >> father >> \verb|#|1; \verb|#|1 >> place of death >> \verb|#|2; final answer: \verb|#|answer \\
    Filled reasoning chain: Strut-Harald >> father >> Gorm the Old; Gorm the Old >> place of death >> Jelling; final answer: Jelling \\
    \bottomrule
  \end{tabular}
  \caption{Examples of in-context learning exemplars used in our method.}
  \label{table:our-method-ic-examples}
\end{table*}

\begin{table*}
  \small
  \centering
  \begin{tabular}{p{.95\textwidth}}
  \toprule
    Question: Were Wessel Dammers and Robert Handcock (Rugby Union) from the same country? \\
    Answer: no \\\\

    Question: Which award the director of film The Blue Umbrella (2005 Film) won? \\
    Answer: National Film Award for Best Music Direction \\

    \ldots \\

    Question: Where was the place of death of Strut-Harald's father? \\
    Answer: Jelling \\
    
    \bottomrule
  \end{tabular}
  \caption{Examples of in-context learning exemplars used in the standard few-shot prompting.}
  \label{table:few-shot-ic-examples}
\end{table*}

\section{In-Context Learning Exemplars}
\label{sec:appendix-ic-examples}

In our experiments on the 2Wiki and MuSiQue datasets, we employ the following settings:
\begin{itemize}
    \item For parsing semi-structured reasoning chains and using the model's parametric memory to fill in masks, we use a 25-shot setting.
    \item For converting masked triplets into single-hop questions and answering these questions, we use a 15-shot setting.
    \item For standard few-shot prompting, we use a 25-shot setting.
\end{itemize}

The experiment in the Bamboogle dataset differs from those mentioned above in that, for parsing semi-structured reasoning chains, using the model's parametric memory to fill in masks, and in standard few-shot prompting, we use a 4-shot setting.

We present examples of the in-context learning exemplars used throughout our method in \Cref{table:our-method-ic-examples} and exemplars used in the standard few-shot prompting experiments in \Cref{table:few-shot-ic-examples}.

\section{Implementation Details}
\label{sec:appendix-knowledge-source-and-useage-details}

\paragraph{Knowledge Sources and Usage Details}
The TKB for 2Wiki consists of 398,354 paragraphs, and for MuSiQue, it contains 139,409 paragraphs. For querying unstructured text knowledge sources, we use the pre-trained Contriever (\verb|Contriever-msmarco|) \cite{izacard2022unsupervised} to retrieve the top $10$ most relevant paragraphs for each single-hop question converted from a structured triplet. We employ the off-the-shelf BLINK entity linking model \cite{wu-etal-2020-scalable} to link the head of the structured triplet to Wikipedia, and use WikiMapper \footnote{\url{https://github.com/jcklie/wikimapper}} to obtain the corresponding Wikidata entry.  For relation linking, we use the \verb|all-MiniLM-L6-v2| model from SentenceTransformers \cite{reimers-gurevych-2019-sentence} to encode the relation string in the structured triplet and all the property labels from Wikidata and calculate the cosine similarity between them to find the most similar Wikidata property. We utilize the results of entity and relation linking to construct simple SPARQL queries and execute them on the Wikidata endpoint to obtain the entities corresponding to the masks.

\paragraph{Model Inference}
We use the vLLM library v0.1.3 \cite{kwon2023efficient} for model inference. All inference is conducted on eight Nvidia A6000 GPUs. The total GPU hours are around 780 hours.

\begin{table}[t]
\small
\centering
\setlength{\tabcolsep}{0.5em}
\begin{tabular}{l c c}
\toprule
Model & MuSiQue Recall@20 & 2Wiki Recall@20 \\
\midrule
Single-round & 0.64 & 0.67 \\
Multi-round & 0.77 & 0.70 \\
\bottomrule
\end{tabular}
\caption{Performance comparison of iterative and one-time retrieval for MuSiQue and 2Wiki datasets.}
\label{table:retrieval-comp}
\end{table}

\section{Baselines Details}
\label{sec:appendix-baselines-details}
We compare our approach to the following state-of-the-art methods:

\paragraph{Standard Few-Shot Prompting}
Standard few-shot prompting  (Standard) \cite{NEURIPS2020_1457c0d6} simply involves prompting the model to generate answers using the few-shot in-context learning exemplars. 
We utilize a 25-shot setting.

\paragraph{CoT Prompting}
CoT prompting \cite{wei2022chain} incorporates intermediate reasoning steps into the few-shot exemplars, guiding the model to generate step-by-step thought processes before producing the final answer.

\paragraph{Self-Ask}
Unlike the CoT approach, where the model generates the entire thinking process and final answer in one pass, Self-Ask \cite{press2023measuring} adopts a multi-round self-questioning method.
Given a multi-hop question, the Self-Ask method prompts the LLM to generate explicit intermediate questions and uses a search engine to find answers to these questions. 
This process continues until the LLM produces the final answer. 

\paragraph{ITER-RETGEN}
ITER-RETGEN \cite{shao2023enhancing}, similar to Self-Ask, answers multi-hop questions through iterative prompting of LLMs and retrieval. 
However, unlike Self-Ask, ITER-RETGEN does not prompt LLMs to generate explicit sub-questions for each iteration. 
Instead, ITER-RETGEN utilizes content generated by the model in the previous iteration along with the given question to perform retrieval. It then prompts LLMs with the question and the retrieved contents using a CoT prompting approach to generate answers.

\paragraph{ReAct}
ReAct prompting \cite{yao2023react} employs a few-shot prompting strategy to prompt the model to engage in multiple rounds of reasoning and action-taking.
"Reasoning" refers to the model generating a thought process based on the input question or observations from the previous actions. 
An "action" refers to task-specific, which may also include utilizing APIs, such as Wikipedia search, to acquire external knowledge.

\paragraph{DSP}
DSP \cite{Khattab2022DemonstrateSearchPredictCR} employs Python programs to outline the process required to answer multi-hop questions.
These programs integrate the use of language models and retrievers to retrieve relevant documents and make predictions for the final answers.
DSP also uses existing labels from the training set and applies semi-supervised learning techniques to bootstrap label training examples' programs, which are then used as exemplars for in-context learning.

\paragraph{Data Synthesis}
Data Synthesis \cite{chen2023efficient} focuses on multi-hop question-answering tasks by establishing a framework that utilizes few-shot in-context learning to prompt LLMs to generate more than one million question-answer pairs. 
The synthesized data is then used to fine-tune the LLMs. 
During the inference stage, Data Synthesis employs a reasoning and action alternation method similar to ReAct, leveraging prompts to guide the fine-tuned LLMs in answering multi-hop questions.

\begin{table}[t]
\small
\centering
\setlength{\tabcolsep}{0.5em}
\begin{tabular}{l c c c}
\toprule
Model & EM & F1 \\
\midrule
Oracle KG + Text + Model & 0.89 & 0.91 \\
Wikidata KG + Text + Model &  0.82 & 0.87 \\
\bottomrule
\end{tabular}
\caption{Comparison of using Wikidata and Oracle KG on 2Wiki with LLAMA2-70b as the base model.}
\label{table:oracle-kg-comp}
\end{table}

\section{Comparison of Using Wikidata and Oracle KG}
\label{sec:appendix-oracle-kg}

We present the performance comparison on 2Wiki between using Wikidata and Oracle KG in \Cref{table:oracle-kg-comp}.

\section{Error Analysis}
\label{sec:appendix-error-analysis}
We show the types of errors annotated and corresponding examples in \Cref{tab:error-analysis}.

\begin{table*}
    \centering
    \small
    \setlength{\tabcolsep}{0.5em}
    \begin{tabular}{c p{0.08\textwidth} c c p{0.2\textwidth} p{0.2\textwidth} p{0.09\textwidth} p{0.08\textwidth}}
    \toprule
    & & \multicolumn{2}{c}{\textbf{Frequency}} \\
    \cmidrule(lr){3-4}
    \textbf{\#} & \textbf{Type} & \textbf{2Wiki} & \textbf{MuSiQue} & \textbf{Question} & \textbf{Reasoning Chain} & \textbf{Prediction} & \textbf{Answer} \\
    \midrule
    1 &
    False Negative 
    & 29 & 11
    & How many natioanl female outdoor track and field championships have been won by the university where Bob Allison was educated?
    & Bob Allison >> university where he was educated >> James Madison University.; James Madison University. >> number of national female outdoor track and field championships won >> 1.
    & 1
    & one
    \\
    \\ \midrule
    2 &
    Knowledge Retrieval
    & 7 & 26
    & Which film has the director who died earlier, It'S Never Too Late (1956 Film) or The Slaughter Rule?
    &  It's Never Too Late (1956 film) >> director >> Michael McCarthy; Michael McCarthy >> date of death >> May 7th 1959; The Slaughter Rule >> director >> Alex Smith; Alex Smith >> date of death >> 2006
    & It's Never Too Late (1956 film)
    & The Slaughter Rule
    \\ \midrule
    3 &
    Reasoning chain
    & 1 & 13
    & What position was it that the person who said man cannot live by bread alone was the Son of God?
    & person who said man cannot live by bread alone >> Son of God >> Jesus.; Jesus. >> position >> king and judge of his kingdom
    & king and judge of his kingdom
    & the Arian position
    \\
    \bottomrule
    \end{tabular}
    \caption{Error categories and examples on 2Wiki and MuSiQue.}
    \label{tab:error-analysis}
\end{table*}

\section{Comparison of Single-round Retrieval and Multi-round Retrieval}
When using unstructured text to fill in reasoning chains, we need to retrieve documents from the knowledge sources. 
There are two options for retrieval: one is to directly use multi-hop questions as the query for the retriever, fetching all relevant content in one shot and then using them to fill the reasoning chain. 
The other option is to iteratively use single-hop questions, converted from triplets, as the retrieval query to fetch relevant documents. 
We adopt the latter approach. In \Cref{table:retrieval-comp}, we report the performance of these two retrieval methods in terms of recall@20. The results show that multi-round retrieval surpasses single-round retrieval, especially in the MuSiQue dataset. This is because the questions in the MuSiQue dataset often require bridging reasoning, where the next hop of reasoning depends on the knowledge from the previous hop, and the questions usually lack the intermediary information that acts as a bridge.

\section{GPT-3.5-Turbo Results}
\begin{table*}
\small
\centering
\begin{tabular}{l c l c c c c}
\toprule
& & & \multicolumn{2}{c}{\textbf{2Wiki}} & \multicolumn{2}{c}{\textbf{MuSiQue}} \\
\cmidrule(lr){4-5}
\cmidrule(lr){6-7}
\textbf{Method} & \textbf{Inference Only} & \textbf{Base Model} & \textbf{EM} & \textbf{F1} & \textbf{EM} & \textbf{F1} \\
\midrule
Standard & \Checkmark & gpt-3.5-tubro   & 0.25 & 0.28 & 0.20 & 0.11 \\
KG + Text + Model & \Checkmark & gpt-3.5-tubro  & 0.73 & 0.80 & 0.49 & 0.33 \\
\bottomrule
\end{tabular}
\caption{Performance comparison of Standard and KG + Text + Model methods using gpt-3.5-turbo on the 2Wiki and MuSiQue datasets.}
\label{table:gpt-3.5-performance-comparison}
\end{table*}

Following previous works \cite{khattab2022demonstrate,trivedi-etal-2023-interleaving,yoran-etal-2023-answering} and considering API costs, we randomly sampled 100 examples from the 2Wiki and MuSiQue datasets for our experiment, utilizing the gpt-3.5-turbo-0613\footnote{\url{https://platform.openai.com/docs/models/gpt-3-5-turbo}}. The experimental setup is identical to the one described in the \Cref{sec:experiments}. The results are shown in \Cref{table:gpt-3.5-performance-comparison} and demonstrate that our approach can significantly improve proprietary LLMs' performance.

\section{Fact Verification Results}

To further validate our method's generalizability, we conduct experiments on a popular fact verification dataset, FEVER \cite{thorne-etal-2018-fever}. Our empirical analysis suggests that fact verification datasets typically rely on specific knowledge sources, which can result in inaccuracies in the ground-truth labels, particularly those annotated with "not enough information" due to a dependency on a single source. To robustly test our approach, which integrates multiple knowledge sources, we randomly select 80 examples from the FEVER test set for manual re-annotation. The experimental setup for our method and the standard few-shot baseline follows the same setup outlined in Section \ref{sec:experiments}. For the CoT baseline, we employ three manually labeled fact verification steps as in-context examples to prompt the model. The results, displayed in \Cref{table:fact-verification-performance}, show that our method outperforms the baseline models, with a 7\% to 14\% relative increase in accuracy rates.

\begin{table*}
\small
\centering
\begin{tabular}{l c c c}
\toprule
\textbf{Method} & \textbf{Inference Only} & \textbf{Base Model} & \textbf{Accuracy} \\
\midrule
Standard & \Checkmark & llama2-70b & 0.69  \\
CoT &  \Checkmark & llama2-70b & 0.74 \\
KG + Text + Model & \Checkmark & llama2-70b & 0.79 \\
\bottomrule
\end{tabular}
\caption{Performance comparison of Standard, CoT, and KG + Text + Model methods on the FEVER dataset.}
\label{table:fact-verification-performance}
\end{table*}

\end{document}